\documentclass[journal]{IEEEtran}
\usepackage[T1]{fontenc}
\usepackage{graphicx}
\usepackage{times}
\usepackage{helvet}
\usepackage{courier}
\usepackage{amsmath}
\usepackage{algorithm}
\usepackage{algorithmic}
\usepackage{csquotes} 
\usepackage{color}
\usepackage{paralist}
\usepackage{amssymb}
\usepackage{indentfirst}
\usepackage{subfigure}
\usepackage{float}
\usepackage{multirow}
\usepackage{cite}
\usepackage{mathrsfs}

\usepackage{mathrsfs} 
\usepackage{amsfonts}
\usepackage{todonotes}
\usepackage{pgfplots} 
\usepackage{epstopdf}
\usepackage{epsfig}
\pgfplotsset{compat=newest}
\usepackage{color}
\usepackage{tabularx}

\definecolor{forestgreen}{RGB}{0,139,69}

\usepackage{xcolor}
\definecolor{citecolor}{HTML}{0071bc}
\usepackage[colorlinks, linkcolor=red,  anchorcolor=blue, citecolor=citecolor]{hyperref} 

\usepackage{xcolor}
\definecolor{SeaGreen4}{RGB}{0,205,102} 
\definecolor{SlateBlue}{RGB}{106,90,205} 
\definecolor{DarkRed}{RGB}{178,34,34}

\usepackage{textcomp,booktabs}
\usepackage{amssymb}
\usepackage{pifont}

\usepackage{makecell}

\usepackage{colortbl}
\definecolor{mygray}{gray}{.9}
\definecolor{mypink}{rgb}{.99,.91,.95}
\definecolor{mycyan}{cmyk}{.3,0,0,0}

\begin{document}

\title{ VFM-Det: Towards High-Performance Vehicle Detection via Large Foundation Models }   



\author{Wentao Wu$^{\dag}$, Fanghua Hong$^{\dag}$, Xiao Wang*, \emph{Member, IEEE}, Chenglong Li, Jin Tang 

\thanks{$\dag$ The first two authors contribute equally to this work.}

\thanks{ $\bullet$ Wentao Wu, Chenglong Li are with Information Materials and Intelligent Sensing Laboratory of Anhui Province, Anhui Provincial Key Laboratory of Multimodal Cognitive Computation, the School of Artificial Intelligence, Anhui University, Hefei 230601, China. (email: wuwentao0708@163.com, lcl1314@foxmail.com)} 

\thanks{$\bullet$ Fanghua Hong is with the School of Electronic and Information Engineering, Anhui University, Hefei 230601, China. (email: fanghuahon@qq.com)} 

\thanks{ $\bullet$  Xiao Wang, Jin Tang are with the School of Computer Science and Technology, Anhui University, Institute of Artificial Intelligence, Hefei Comprehensive National Science Center, Hefei 230601, China. (email: \{xiaowang, tangjin\}@ahu.edu.cn)} 

\thanks{* Corresponding author: Xiao Wang} 
}

\markboth{IEEE Transactions on Intelligent Transportation Systems, 2024}   
{Shell \MakeLowercase{\textit{et al.}}: Bare Demo of IEEEtran.cls for IEEE Journals}

\maketitle

\begin{abstract}
Existing vehicle detectors are usually obtained by training a typical detector (e.g., YOLO, RCNN, DETR series) on vehicle images based on a pre-trained backbone (e.g., ResNet, ViT). Some researchers also exploit and enhance the detection performance using pre-trained large foundation models. However, we think these detectors may only get sub-optimal results because the large models they use are not specifically designed for vehicles. In addition, their results heavily rely on visual features, and seldom of they consider the alignment between the vehicle's semantic information and visual representations. In this work, we propose a new vehicle detection paradigm based on a pre-trained foundation vehicle model (VehicleMAE) and a large language model (T5), termed VFM-Det. It follows the region proposal-based detection framework and the features of each proposal can be enhanced using VehicleMAE. More importantly, we propose a new VAtt2Vec module that predicts the vehicle semantic attributes of these proposals and transforms them into feature vectors to enhance the vision features via contrastive learning. Extensive experiments on three vehicle detection benchmark datasets thoroughly proved the effectiveness of our vehicle detector. Specifically, our model improves the baseline approach by $+5.1\%$, $+6.2\%$ on the $AP_{0.5}$, $AP_{0.75}$ metrics, respectively, on the Cityscapes dataset. The source code of this work will be released at \url{https://github.com/Event-AHU/VFM-Det}. 
\end{abstract}


\IEEEpeerreviewmaketitle

\section{Introduction} \label{sec::introduction}

\IEEEPARstart{V}{ehicle} detection is the premise of fine-grained vehicle analysis and plays an important role in intelligent video surveillance. Nowadays, many object detectors can be adopted for vehicle detection, such as YOLO series~\cite{redmon2016you, redmon2017yolo9000, farhadi2018yolov3, bochkovskiy2020yolov4}, R-CNN series~\cite{girshick2015region, ren2015faster, he2017mask, girshick2014rich}, DETR series~\cite{carion2020end, zhu2020deformable, zhang2022dino,li2022exploring}, and other detectors~\cite{chen2023edge, zhu2023multi, jia2023tfgnet, zhang2024night}. Although good performance can already be achieved using these models, however, the detection performance in challenging scenarios is still unsatisfactory. Many researchers employ multi-modal data to address these issues, however, the popularity of multi-modal devices is still low, which limits the scope of use of their methods. 

\begin{figure}
    \centering
    \includegraphics[width=3.3in]{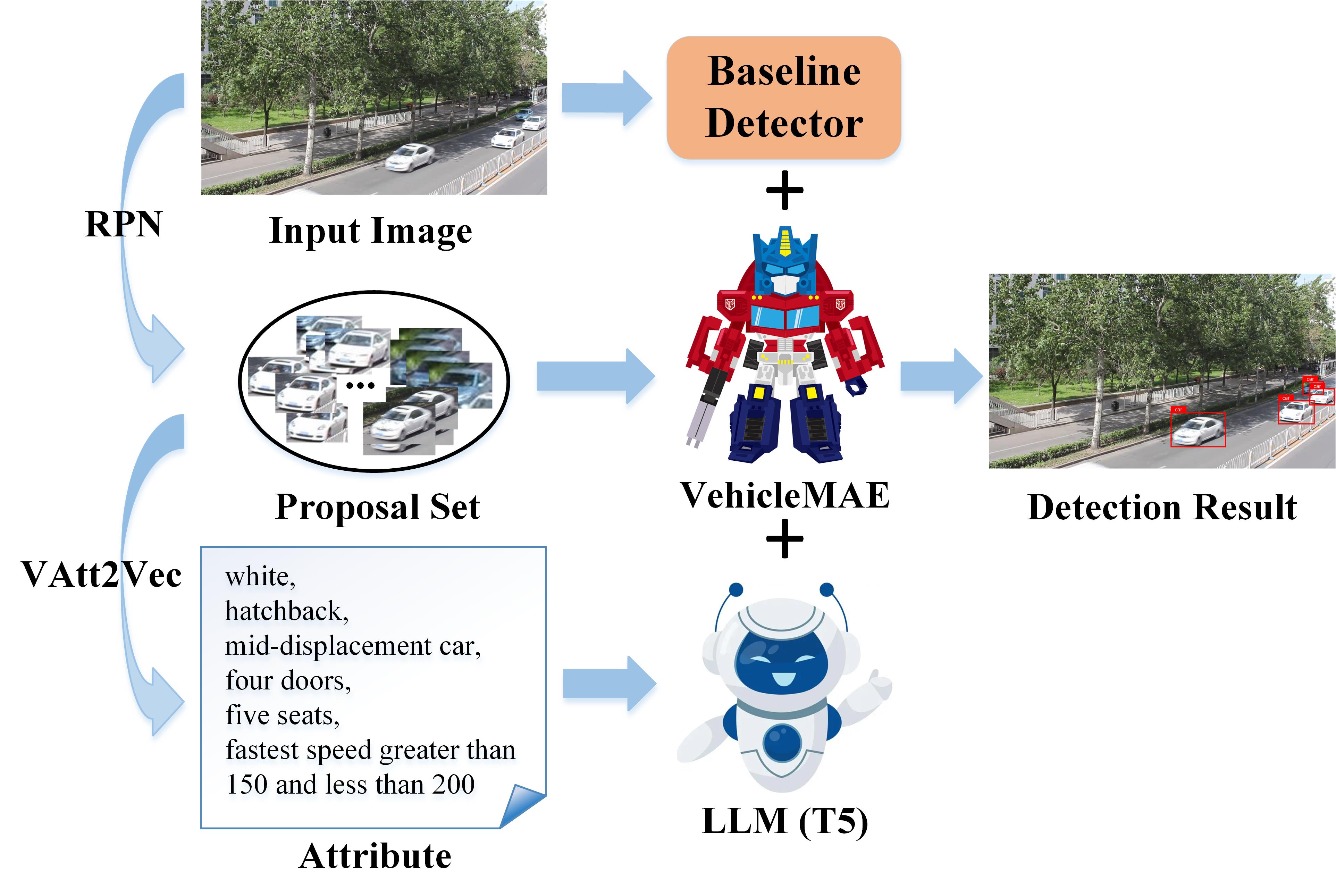}
    \caption{Illustration of our proposed the Large Vision Model (VehicleMAE~\cite{wang2023structural}) and Large Language Model (T5~\cite{raffel2020exploring}) based vehicle detection.} 
    \label{fig:FirstIMG} 
\end{figure}

Recently, pre-training and fine-tuning techniques have become very popular in the artificial intelligence (AI) community. The researchers usually first pre-train a large backbone network (for example, BERT~\cite{kenton2019bert}, GPT series~\cite{radford2018GPT1, radford2019GPT2, brown2020GPT3, openai2023gpt4}, ViT~\cite{dosovitskiy2020image}, CLIP~\cite{radford2021learning}) in a self-supervised or unsupervised way, then, adapt it to various downstream tasks using Parameter Efficient Fine Tuning (PEFT) strategies (e.g., prompt tuning~\cite{jia2022VPT}, adapter~\cite{houlsby2019adapter}, side tuning~\cite{zhang2020sidetuning, wang2024STsideTuning}, etc). Inspired by these works, some researchers attempt to tune these pre-trained big models for object detection. Specifically, Li et al.~\cite{li2022exploring} develop a backbone network for the object detection task using the plain and non-hierarchical vision Transformer. Lin et al.~\cite{lin2024LargeUniDet} address the multi-domain universal detection problem using a pre-trained large vision model. Fang et al.~\cite{fang2023MIMDet} used pre-trained vanilla ViT network for object detection and instance segmentation. They achieve better results even compared with strong hierarchical architectures, such as Swin-Transformer~\cite{liu2021swintransformer}, MViTv2~\cite{li2022mvitv2} and ConvNeXt~\cite{woo2023convnext}.

Although a better detection performance can be achieved, however, these models may be still limited by the following issues: 
\emph{1). Generalized Large Model vs. Large Vehicle Model}: 
Current pre-trained big vision model based object detectors are proposed for universal or general object detection which usually adopts the generalized pre-trained large backbone for their work. However, these large models may only achieve sub-optimal results for the vehicle detection problem which is a specific downstream task. 
\emph{2). Semantic Gap Problem}: 
Existing vehicle detection algorithms mainly rely on the feature representations obtained using the pure pre-trained large vision models, but ignore the semantic gap between the semantic label information of the target object (e.g., the vehicle) and the visual representations. 
Given the two issues above, it is natural to raise the following question: \textit{"How can we achieve high-performance vehicle detection using pre-trained models and fully bridge the aforementioned semantic gaps between the vision features and semantic categories?"}

In this work, we propose a new vehicle detection framework by adapting the pre-trained large vehicle foundation model VehicleMAE~\cite{wang2023structural} and a large language model (T5~\cite{raffel2020exploring} adopted in this work) into the Mask RNN detector, as illustrated in Fig.~\ref{fig:FirstIMG}. More in detail, on the basis of region proposal-based detection framework, our model takes the raw image as the input and adopts a backbone network (e.g., ResNet~\cite{he2016resnet}) for feature extraction. After that, it adopts the Region Proposal Network (RPN) to get the features of interest. A detection head is introduced for location regression and label classification. In this work, we propose to enhance the baseline detector from the following two aspects. First, the large vision model VehicleMAE which is specifically pre-trained on massive vehicle images is adopted to enhance the feature representations of each proposal. Second, we predict the semantic cues of each proposal using an attribute estimate network which takes the vision features of each proposal and all the attributes as the input. We transform the estimated attributes into a unified semantic representation using GRU network~\cite{cho2014GRU} and bridge the aforementioned semantic gaps by aligning the vision features and unified semantic representation based on contrastive learning. The experiments thoroughly proved the effectiveness of our proposed strategy for the vehicle detection.

To sum up, the key contributions of this paper can be summarized as the following three aspects: 

1). We propose a novel vehicle detection framework, termed \textit{VFM-Det}, by adapting the pre-trained vehicle large model VehicleMAE and successfully improving the final results significantly. 
    
2). We propose a new semantic gap reduction module, termed \textit{VAtt2Vec}, for vehicle detection by learning the representations of vehicle attributes and aligning its vision features using contrastive learning. 

3). Extensive experiments on three large-scale vehicle detection benchmark datasets  thoroughly proved the effectiveness of our proposed vehicle detection framework.

\textit{The following of this paper is organized as follows:} In section~\ref{sec::relatedWorks}, we mainly introduce the related works about the pre-trained models, object detection task, and attribute representation learning. In section~\ref{sec::method}, we concentrate on outlining our proposed framework, more in detail, we first review the scheme of the proposal-based detection framework and then give a review of our proposed new vehicle detection model. After that, we focus on the pre-trained VehicleMAE-based perceptron, vehicle attribute representation learning, and loss functions used for training. In section~\ref{sec::experiments}, we introduce our experiments by describing the datasets and evaluation metrics, implementation details, comparisons with other detectors, and related analysis. We also summarize the limitations of our model and propose possible improvements as our future works in section~\ref{subsec:limitation}, and conclude this work in section~\ref{sec::Conclusion}.

\section{Related Works} \label{sec::relatedWorks}
In this section, we will introduce the works most related to ours, including Object Detection, Pre-trained Big Models, and Attribute Representation Learning~\cite{wang2022PARsurvey}\footnote{\url{github.com/wangxiao5791509/Pedestrian-Attribute-Recognition-Paper-List}}.

\subsection{Object Detection} 
The three main categories of deep learning-based object detection are one-stage detectors~\cite{redmon2016you, lin2017focal}, two-stage detectors~\cite{girshick2015region}, and Transformer-based detectors~\cite{carion2020end, li2022exploring}. 
Among them, the one-stage detectors based on sliding windows directly classify and locate semantic targets through dense sampling, avoiding the step of screening potential object regions in the image. YOLO~\cite{redmon2016you} adopts a direct regression approach to obtain detection results, effectively improving the detection speed. RetinaNET~\cite{lin2017focal} proposes a focal loss function, which handles the foreground-background imbalanced problem well. YOLO-V3~\cite{farhadi2018yolov3} adopts multi-scale feature map extraction, improving the detection effect of the YOLO series on small targets significantly. After that, a series of follow-up works~\cite{bochkovskiy2020yolov4,wong2019yolo,li2022yolov6} are proposed to further improve the YOLO detector. 
Two-stage object detection first extracts the proposal from the image, and then makes secondary modifications grounded in the proposal to obtain the detection results. R-cnn~\cite{girshick2015region} first attempts to address the detection task using neural networks and improve the overall performance significantly. Faster RCNN~\cite{ren2015faster} proposes an RPN module to generate candidate boxes that solve the issues caused by selective search well. Mask RCNN~\cite{he2017mask} introduces the RoI align layer instead of RoI pooling operator to avoid pixel-level dislocation caused by spatial quantization.

Along with the outstanding performance of Transformer networks in many domains, some researchers have begun to think about the integration of Transformer and object detection. To be specific, the detectors~\cite{qiao2021detectors} based on Transformer, DETR~\cite{carion2020end} first introduce Transformer into object detection. Zhu et al.~\cite{zhu2020deformable} proposed the deformable attention module, which improved the training speed and detection performance on small objects of the original DETR algorithm. DINO~\cite{zhang2022dino} adopts contrastive denoising training and mixed query selection methods, further enhancing the performance of the DETR model. VITDET~\cite{li2022exploring} uses SFP instead of a feature pyramid structure, eliminating hierarchical constraints of backbone networks. 
Different from these works, in this paper, we propose to enhance vehicle detection using pre-trained large vision and language models and achieve a higher performance than our baseline significantly.

\subsection{Pre-trained Big Models} 
Self-supervised/unsupervised pre-training is currently the focus of research. Currently, there are two mainstream self-/unsupervised pre-trained algorithms: contrastive learning and reconstruction-based pre-trained. 
Specifically, contrastive learning methods aim to train the network to distinguish whether given input pairs are aligned. For single-modal input pre-training, SimCLR~\cite{chen2020simple}, MoCo~\cite{he2020momentum}, etc., generate a set of similar samples through data augmentation strategies. BYOL~\cite{grill2020bootstrap} learns to encode the similarity of positive samples to construct representations. For multi-modal input pre-training, CLIP~\cite{radford2021learning} and ALIGN~\cite{jia2021scaling} pre-trained on image-text pairs, while VideoCLIP~\cite{xu2021videoclip} extends it to video-text pairs. Reconstruction-based methods attempt to train the network to reconstruct the masked parts of the input to learn representations. BERT~\cite{kenton2019bert} and its extensions use a bidirectional transformer and present few-shot learning capabilities from masked language modeling. MAE~\cite{he2022mae} proposes masked autoencoders that learn visual representations by simply reconstructing pixels. VideoMAE~\cite{tong2022videomae} and VideoMAE-v2~\cite{wang2023videomae} further apply this to the video domain, while MultiMAE~\cite{bachmann2022multimae} extends the input to multi-modal, learning richer geometric and semantic information. There are also some pre-training methods designed for specific targets, such as HCMoCo~\cite{hong2022versatile}, HumanBench~\cite{tang2023humanbench},  SOLIDER~\cite{chen2023beyond}, etc., which focus on human-centered pre-training, while VehicleMAE~\cite{wang2023structural} focuses on vehicle-based pre-training. Motivated by the effectiveness of these works, we utilize the pre-trained large models in our study to improve the vehicle detector even more.

\subsection{Attribute Representation Learning} 
The attributes of the target object can fully reflect the key cues, e.g., the texture, color, shape, etc, and can also be obtained easily. Therefore, many researchers exploit the semantic attributes for their tasks, for example, the re-identification. 
Lin~\cite{lin2019improving} et al. proposed a method that utilizes attribute information to enhance re-identification performance. 
Zhang~\cite{zhang2020person} et al. proposed an attribute attentional block to address the interference of noisy attributes on the model. By employing reinforcement learning, they aim to remove noisy attributes and improve the robustness of the model. 
Jeong~\cite{jeong2021asmr} et al. introduced a loss function for learning cross-modal embeddings. This approach treats a set of attributes as a class of individuals sharing the same characteristics, thereby reducing the modality gap between attributes and images. 
Li~\cite{li2023clip} et al. fully leveraged the cross-modal description capabilities of CLIP by providing a set of learnable text tokens for each individual ID. These text tokens are fed into the text encoder to generate ambiguous descriptions, which facilitate better visual representation. 
Additionally, Zhai~\cite{zhai2023multi} et al. introduced fine-grained attribute descriptions as prompt information to provide richer semantic information for re-identification tasks. 
Zhang et al.~\cite{zhang2020AttPersonDet} 
designed an attribute-based non-maximum suppression algorithm to address the issue of missed pedestrian detection in crowded scenes. This algorithm captures high-level semantic differences between crowds by modeling pedestrian attribute information.
Tian et al.~\cite{tian2015personDetAttributes} exploit the semantic tasks to assist pedestrian detection, including pedestrian and scene attributes to help classify the positive with hard negative samples. 
In this work, we estimate the attributes of extracted proposals and learn a unified attribute representation for contrastive learning between the semantic cues and vision features. Our experiments thoroughly proved the effectiveness of our proposed strategy for high-performance vehicle detection.

\begin{figure*}
\centering
\includegraphics[width=7in]{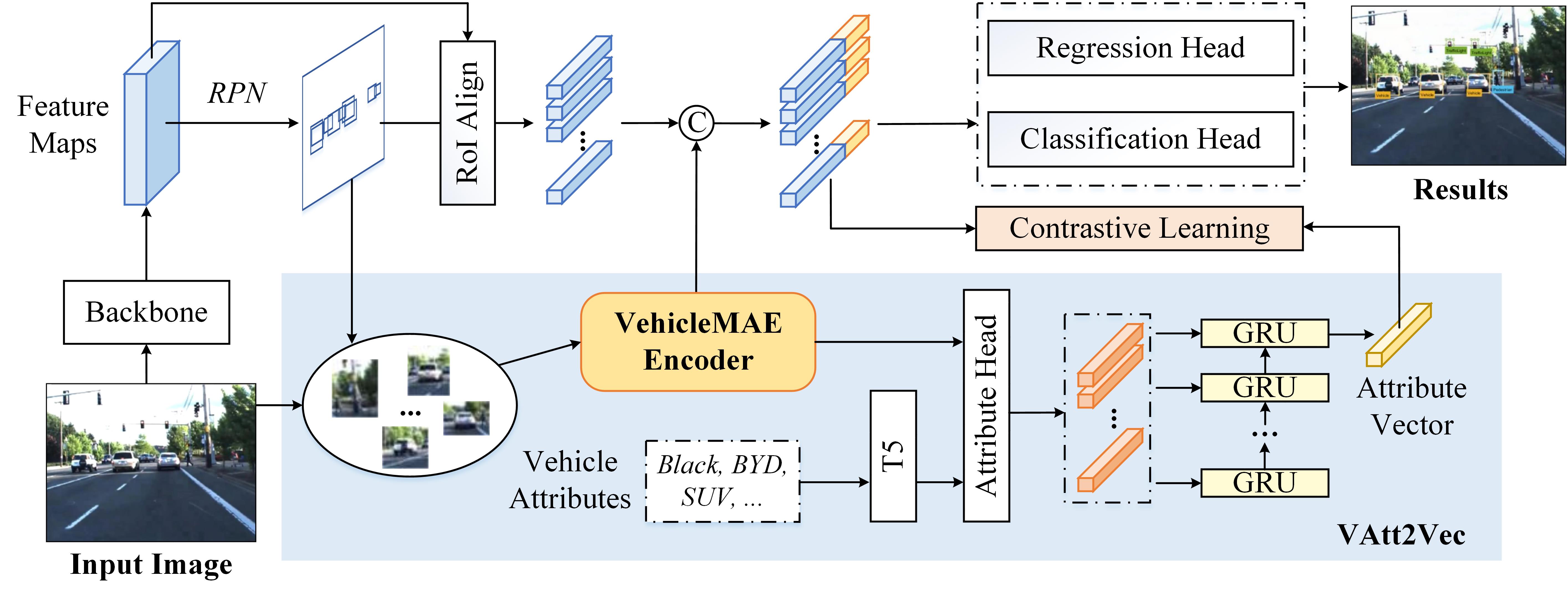}
\caption{ An overview of our proposed VFM-Det framework which exploits vision feature enhancement and semantic-vision alignment for high-performance vehicle detection using pre-trained foundation models. On the basis of the region proposal-based detectors, we introduce the VehicleMAE encoder to enhance the vision features. Meanwhile, we predict the vehicle attributes and learn a unified attribute representation via \textit{VAtt2Vec}, to achieve alignment between the vision and semantic high-level features. More details about the VAtt2Vec module can be found in Fig.~\ref{fig:VAtt2Vec}. Extensive experiments on three vehicle detection benchmark datasets thoroughly proved the effectiveness of our proposed strategies. 
} 
\label{fig:framework} 
\end{figure*}

\section{Methodology}  \label{sec::method} 
In this section, we will first provide an overview of our proposed VFM-Det framework, and then, we review the region proposal-based detection framework. After that, we will present the detailed network architectures, focusing on the VehicleMAE~\cite{wang2023structural} based Perceptron, Contrastive Learning between vision features and unified attribute representations. Finally, we will introduce the detection head and loss functions used for the optimization of our whole framework.

\subsection{Overview} 
In this work, based on the aforementioned region proposal-based object detection framework, we propose applying pre-trained foundation models to the vehicle detection tasks, as shown in Fig.~\ref{fig:framework}. Specifically, we first feed the input image into the ResNet-50 backbone network to get the feature maps. Then, the RPN generates a group of candidate bounding boxes, and their features can be obtained via the RoI Align layer. More importantly, we crop the proposals out and feed them into the pre-trained vehicle foundation model VehicleMAE to extract more fine-grained vehicle-specific features. These features are concatenated with the features output by RoI Align along the channel dimension and then fed into the detection head and classification head to obtain the target's location and category. 
In addition, we also propose a new \textit{VAtt2Vec} module to bridge the gaps between the vision features used for vehicle detection and semantic labels that can describe the vehicles in high-level semantic features. To be specific, this module takes the vision features of each proposal and all the defined vehicle attributes as the input and predicts the attributes using an attribute head. Note that, the given attributes are encoded using a large language model T5~\cite{raffel2020exploring}. The predicted attributes are further fused and transformed into a unified feature representation via a GRU~\cite{cho2014GRU} module. The visual feature and unified semantic attribute feature are used for contrastive learning. Extensive experiments on three vehicle detection benchmark datasets demonstrate that the pre-trained large vision and language models improve vehicle detection performance significantly.

\subsection{Review: Region Proposal-based Detection} 
The key ideas of the region proposal-based detection framework can be divided into two stages, i.e., the \textit{candidate region generation} and \textit{classification}. To be specific, we first generate a series of candidate regions using an efficient algorithm, which contains positive and negative objects. Then, a \textit{classification head} and a \textit{bounding box regression head} are adopted for each proposal to determine target object is present or not, and also predict its locations and scales more accurately. One can find that the high-quality candidate region generation is a key procedure in the region proposal-based detection framework and the widely used modules including selective search, edge boxes, RPN, etc. Selective Search produces candidate regions by dividing the image into multiple regions and evaluating the properties of each region; edge boxes generate candidate regions guided by image edge information; RPN predicts the location and size of candidate regions by training neural networks. The representative region proposal-based detectors contains RCNN~\cite{girshick2015region}, Faster R-CNN~\cite{ren2015faster}, Mask R-CNN~\cite{he2017mask}, etc. In this work, we integrate our proposed strategies into the Mask R-CNN detector to validate their effectiveness. The subsequent subsection will provide further information, correspondingly.

\subsection{VehicleMAE based Perceptron} 
Based on the Mask R-CNN~\cite{girshick2014rich}, we propose to enhance the detection performance on the vehicles using a specifically pre-trained foundation vision model VehicleMAE~\cite{wang2023structural}. Specifically, given the input image $I$ and the candidate region information $C$ generated by the RPN module, for each input image, the RPN generates 512 candidate regions. We first crop all proposal images from the input image based on the candidate region information $C$, and resize them into $224 \times 224$, and obtain the image set  $I^{pro} \in \mathbb{R}^{224 \times 224 \times 3}$. For each image in the image set $I^{pro}$, we divide it into \textit{196} non-overlapping patches, then, project them into token embeddings $P^{pat}_{emb} \in \mathbb{R}^{1 \times 768}, j \in \{1, 2, ..., 196\}$, using a convolutional layer (kernels size is $16 \times 16$). After integrated \textit{CLS}-token, we have obtained the feature $F^{CLS}_{emb} \in \mathbb{R}^{197 \times 768}$. We introduce position encoding $Z^{pos} \in \mathbb{R}^{197 \times 768}$ to encode the spatial coordinates of input tokens. More in detail, the position encoding is randomly initialized and added with the token embedding, i.e., we have $F^{pos}_{emb} = Z^{pos} + F^{CLS}_{emb}$. Significantly, we fixed the parameters of the pre-trained VehicleMAE network and introduced learnable tokens $K \in \mathbb{R}^{8 \times 768}$ to achieve more efficient fine-tuning. These learnable tokens are placed between the \textit{CLS}-token and patch-token to obtain feature $F_{emb} \in \mathbb{R}^{205 \times 768}$.

\begin{figure*}
    \centering
    \includegraphics[width=7in]{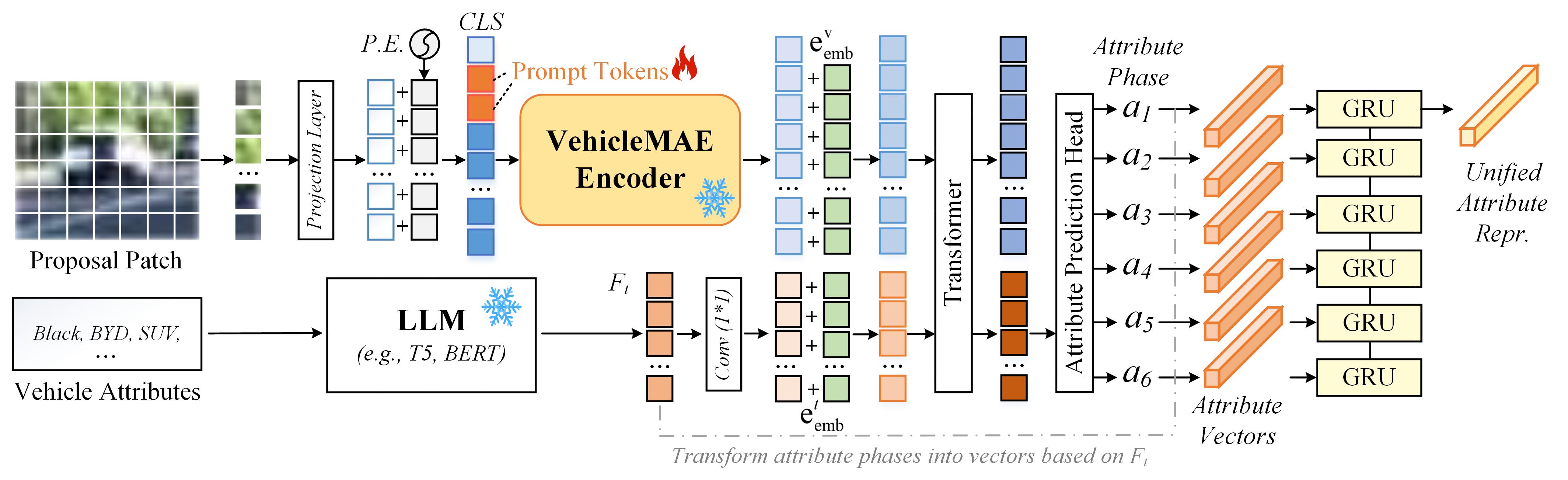}
    \caption{The detailed architecture of our proposed VAtt2Vec module.}
    \label{fig:VAtt2Vec}
\end{figure*}

After obtaining the input vision embeddings $P_{emb}$, we input them into the VehicleMAE encoder (i.e., ViT-B/16) which contains 12 Transformer blocks. 
The output of the foundation vision model VehicleMAE is $\tilde{F} \in \mathbb{R}^{205 \times 768}$. To align with the dimension of RoI Align output features from the ResNet50 backbone network $F_{roi} \in \mathbb{R}^{256 \times 16 \times 16}$, we use a 256-dimensional linear projection layer to project the output of VehicleMAE encoder $\tilde{F}$ to $\bar{F} \in \mathbb{R}^{205 \times 16 \times 16}$. Finally, we concatenate the RoI Align output features $F_{roi}$ and $\bar{F}$ along the channel dimension to obtain the visual feature $F_{v} \in \mathbb{R}^{461 \times 16 \times 16}$ for vehicle detection.

By introducing the pre-trained foundation vehicle-specific vision model, the detection performance can be improved. However, this detector still ignores the alignment between the vision features and semantic high-level features, which may still bring us a sub-optimal result. In the following sub-section, we will exploit the vehicle attributes guided visual feature learning to further improve overall performance.

\subsection{Vehicle Attribute Representation Learning} 

\begin{figure}
    \centering
    \includegraphics[width=2.7in]{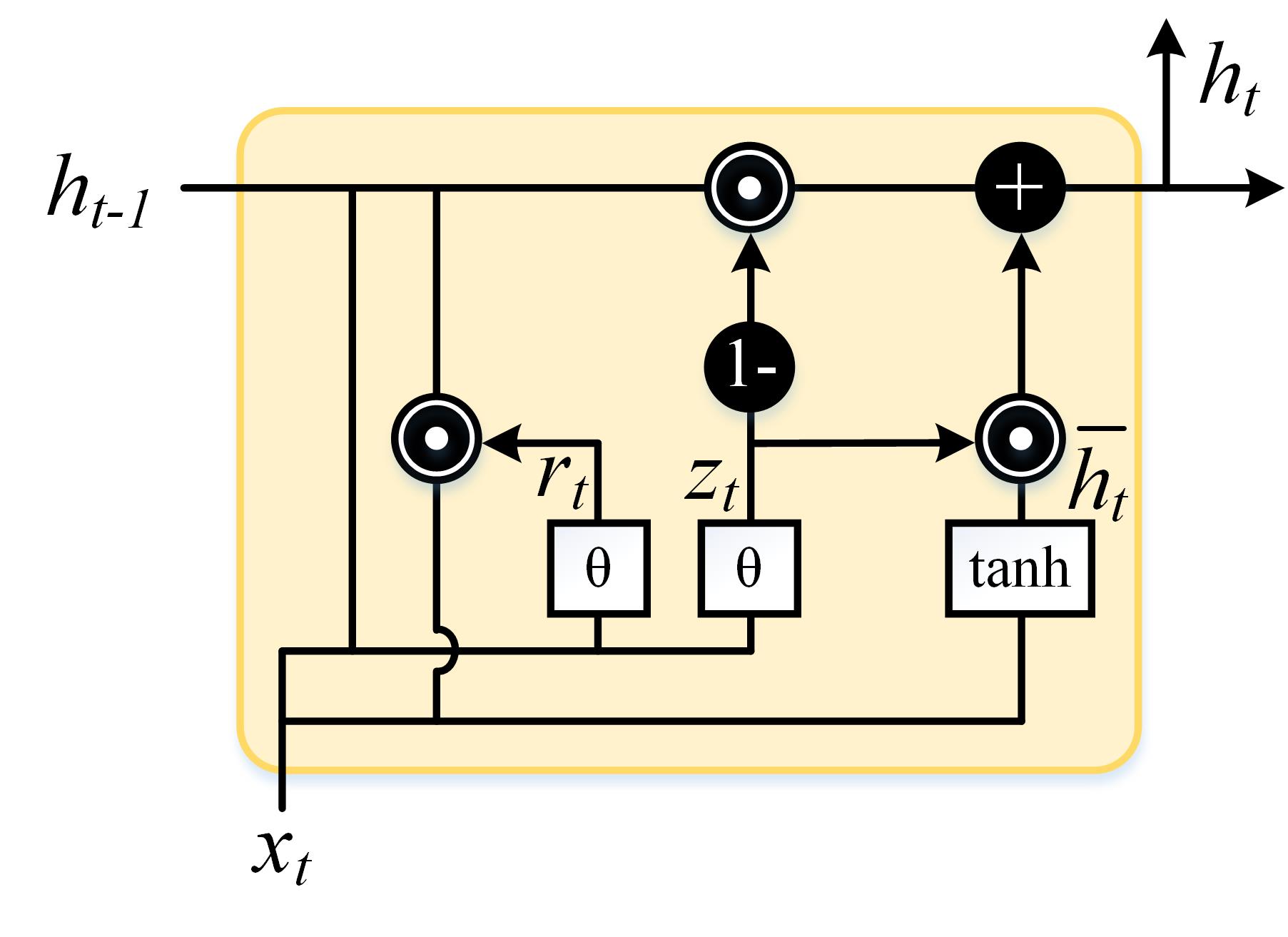}
    \caption{Structural diagram of GRU module.} 
    \label{fig:GRU} 
\end{figure}

To integrate the vehicle attributes into our detection framework, in this work, we define a total of 47 vehicle attribute tags which can be categorized into six groups based on attribute labels from the CompCars~\cite{yang2015large} dataset. More in detail, the six groups are \textit{color}, \textit{number of doors}, \textit{model}, \textit{displacement},  \textit{top speed}, and \textit{number of seats}. We utilize the pre-trained large language model T5~\cite{raffel2020exploring} to obtain the text embeddings $t^{i}_{emb} \in \mathbb{R}^{1 \times 768}, i \in \{1, 2, ..., 47\}$, for these 47 vehicle attribute tags. By concatenating all the text embeddings, we obtain the text feature $F_{t} \in \mathbb{R}^{47 \times 768}$ and feed them into the attribute prediction head along with the visual feature $F_{ve} = \tilde{F}$ output by the encoder of VehicleMAE.

In the attribute prediction head, we first project the text features $F_{t}$ into the same dimension through $1 \times 1$ convolution layer with visual features $F_{ve}$. Here, the feature dimension output by the T5 model is identical to the visual feature dimension, therefore there is no need for projection. After that, we introduce learnable visual embedding $e^{v}_{emb} \in \mathbb{R}^{1 \times 768}$ and text tokens $e^{t}_{emb} \in \mathbb{R}^{1 \times 768}$, to sum with the corresponding features to maintain modality information, respectively. The produced visual features $\tilde{F}_{ve}$ and text features $\tilde{F}_{t}$ are then concatenated to get the feature $F$:
\begin{equation}
F = [\tilde{F}_{t} + e^{t}_{emb}, \tilde{F}_{ve} + e^{v}_{emb}]
\end{equation}
where $[,]$ denotes the concatenate operation.
We input the features $F$ into a layer of Transformer fusion blocks and ultimately output the features as $\bar{F}$. The predicted attributes consist of 47 embeddings corresponding to the text feature $F_{t}$ in the output feature. To avoid the influence of attribute order, we obtain the prediction probabilities $m_{i}, i \in \{1, 2, ..., 47\}$ for each attribute through a group of independent feed-forward networks. 



Based on the predicted attributes, we select the attribute label with the highest probability from each attribute group and extract the corresponding attribute embeddings $a_{i} \in \mathbb{R}^{1 \times 768}, i \in \{1, 2, ..., 6\}$  from the textual features $F_{t}$. This allows us to obtain the attribute embeddings group $A = [a_{1}, a_{2}, ..., a_{6}]$  specific to the given proposal. To obtain a unified semantic attribute representation, we employ a GRU~\cite{cho2014GRU} module to fuse the attribute embeddings, as shown in Fig.~\ref{fig:GRU}, which is effective in capturing dependencies within long sequences. 
By treating each attribute embedding $a_{i}$ as an element in the input sequence, we utilize the GRU's processing capabilities to effectively integrate these features into a unified representation $V_{a} \in \mathbb{R}^{1 \times 256}$. 
An illustration of our proposed VAtt2Vec can be found in Fig.~\ref{fig:VAtt2Vec}.

\subsection{Detection Head and Loss Function} 

For the detection head, we keep the same structure as the original Mask R-CNN~\cite{girshick2014rich}. We input the concatenated visual features $F_{v}$ into the detection head, where they are first projected through two MLP networks and then fed into two fully connected layers. They are respectively used to predict the class score for each proposal and the bounding box regression parameters corresponding to each candidate region, which are further utilized to compute the final box coordinates.

Based on Mask R-CNN~\cite{girshick2014rich}, we introduce a new loss function for contrastive learning between vision features and vehicle semantic attributes. Firstly, we normalize the image $F_{v}$ and text features $V_{a}$ using the $L_2$ norm, and then compute the cosine embedding loss between the normalized features. The following is the formula:
\begin{equation}
L_{va} = \frac {1} {N} \sum^N_{i=1} CEL(\frac {F^v_i} {||F^v_i||_{2}}, \frac {V^a_i} {||V^a_i||_{2}})  
\end{equation}
where $CEL$ stands for Cosine Embedding Loss, $N$ represents the batch size, and $||*||_2$ refers to the $L_{2}$ norm. The final expression for the total loss function $ \mathcal{L} = L_{cls} + L_{reg} + L_{va} $. The bounding box regression loss $L_{reg}$ and classification loss $L_{cls}$ are identical to those defined in Mask R-CNN~\cite{girshick2014rich}. For further information, we advise readers to review their work.

\section{Experiments}  \label{sec::experiments} 
In Section~\ref{DEdatasets}, we will introduce the three vehicle object detection datasets and evaluation metrics. Then, the implementation details are introduced in Section~\ref{impleDetails}. In Section~\ref{citySOTA}, we will compare our method with the state-of-the-art (SOTA) detector and pre-trained models, respectively. After that, we study the effectiveness of each key component in our VFM-Det model in Section~\ref{ablationStudy}. We also give the analysis of tradeoff parameters. In Section~\ref{visualization} and \ref{subsec:limitation} respectively introduced the visualization and limitation analysis.

\begin{figure*}
\centering
\includegraphics[width=7in]{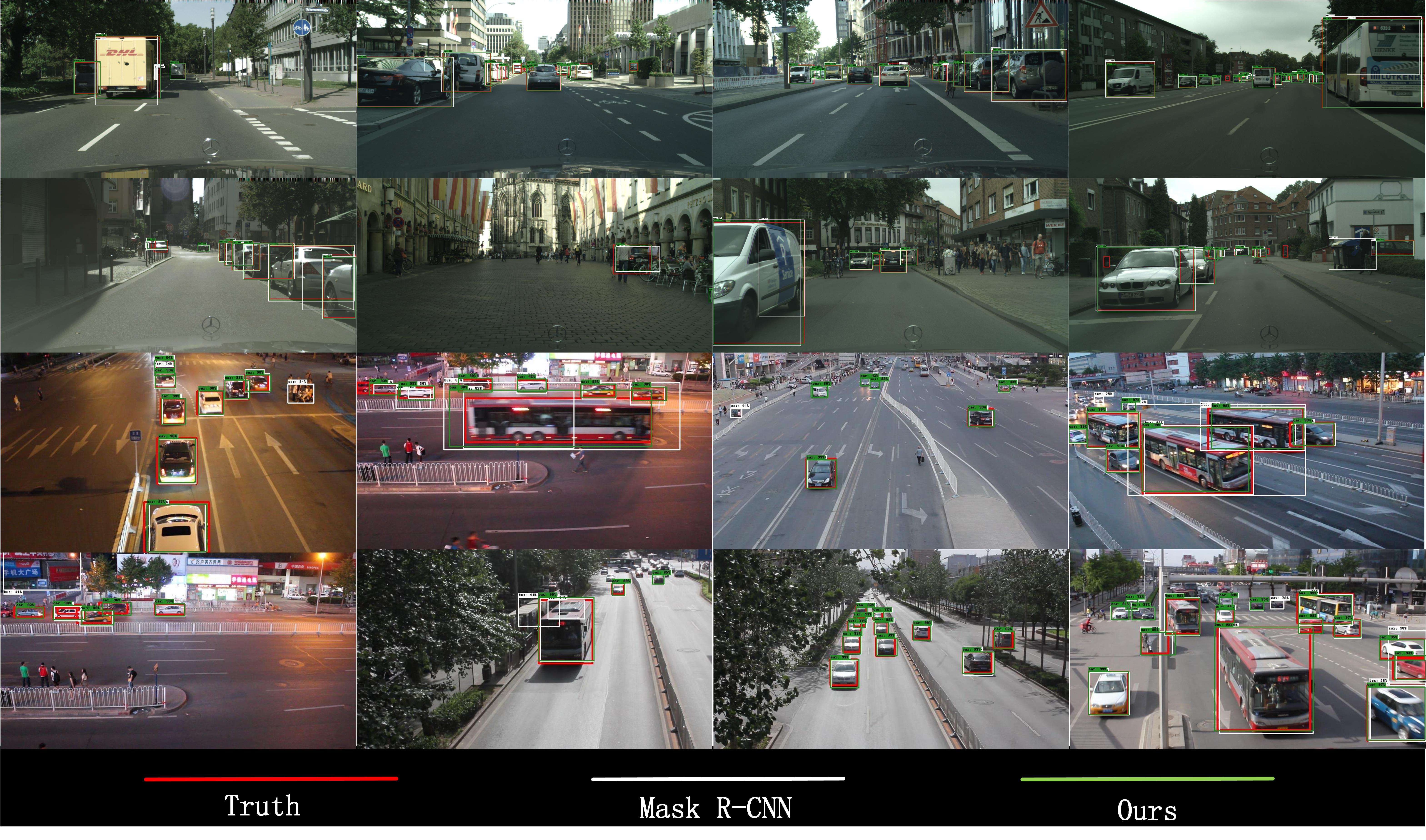}
\caption{Visualization of the detection results from Mask R-CNN and VFM-Det.}
\label{fig:Visualization1} 
\end{figure*}

\subsection{Datasets and Evaluation Metrics} \label{DEdatasets}

\noindent 
$\bullet$ \textbf{Datasets. }  
We have demonstrated the effectiveness and generalization of our proposed VFM-Det on three vehicle target detection datasets. Below is a brief introduction to these datasets. We conducted validation on three vehicle detection datasets, including the Cityscapes~\cite{cordts2016cityscapes} dataset, the UA-DETRAC~\cite{wen2020ua} dataset, and the COCO2017~\cite{lin2014microsoft} dataset.

\noindent 
\textbf{Cityscapes: } 
This is a dataset of urban street scenes. It contains 3,257 high-resolution images from 50 cities in Germany. The dataset covers street scenes under different lighting conditions such as morning, daytime, and nighttime. Each image has a resolution of $2048 \times 1024$ and is annotated for objects such as buildings, roads, pedestrians, and vehicles. In our experiments, we selected four categories related to the vehicle target, i.e., \textit{car}, \textit{bus}, \textit{truck}, and \textit{caravan}. Thus, the dataset contains 2846 training images and 481 testing images.

\noindent 
\textbf{UA-DETRAC: } 
This is a multi-target detection dataset for urban road scenes. For a more detailed introduction, please refer to ~\cite{wen2020ua}. To facilitate training and testing, we constructed a dataset for vehicle object detection by sampling images every 10 frames from the video dataset, it contains 8,178 training images and 5,608 testing images. The dataset includes three vehicle categories, including \textit{car}, \textit{bus}, and \textit{vans}.

\noindent 
\textbf{COCO2017: }
This is an object detection dataset. For a more detailed introduction, please refer to ~\cite{lin2014microsoft}. We select images of the vehicle categories for training using the official API. After processing, the dataset contains 16270 training images and 707 testing images. In our training phase, we selected three vehicle target categories, i.e., \textit{car}, \textit{bus}, and \textit{truck}.

\noindent 
$\bullet$ \textbf{Evaluation Metrics. }
In the work, we have selected three commonly used metrics for object detection, namely $AP_{[0.5:0.95]}$, $AP_{0.5}$, and $AP_{0.75}$. Specifically, $AP_{[0.5:0.95]}$ refers to the calculation of $AP$ values at various $IoU$ (Intersection over Union) thresholds ranging from 0.5 to 0.95 with a step size of 0.05, followed by computing the average of these values. $AP_{0.5}$ represents the $AP$ value when the $IoU$ is set to 0.5, while $AP_{0.75}$ corresponds to the $AP$ value when the $IoU$ is set to 0.75. 
To be specific, the formulaic expression of $AP$ can be written as: 
\begin{flalign}  
& AP = \frac{1}{M} \sum_{j=1}^{M}(Precision_{j}), \\ 
& Precision = TP/(TP + FP),  
\end{flalign} 
where $TP$ and $FP$ represent the number of true positives, and false positives, respectively.

\begin{table*}[!htp]
\centering 
\caption{Experimental results of VFM-Det and other detection algorithms on three object detection datasets: cityscapes, coco, and ua-detrac.} 
\label{detection} 
\vspace{5pt} 
\resizebox{\textwidth}{!}{
\begin{tabular}{c|c|ccc|ccc|ccc|c}  
\hline \toprule [0.5 pt]
\multirow{2}{*}{\raggedright \textbf{Method}} &\multirow{2}{*}{\raggedright \textbf{Backbone}}
&\multicolumn{3}{c|}{\textbf{Cityscapes}}  &\multicolumn{3}{c|}{\textbf{COCO}} &\multicolumn{3}{c|}{\textbf{UA-DETRAC}}
&\multirow{2}{*}{\raggedright \textbf{Params(M)}}\\
& &$AP_{[0.5:0.95]}$  &$AP_{0.5}$ & $AP_{0.75}$ &$AP_{[0.5:0.95]}$  &$AP_{0.5}$ & $AP_{0.75}$ 
&$AP_{[0.5:0.95]}$  &$AP_{0.5}$ & $AP_{0.75}$ \\
\hline 
Faster R-CNN &ResNet50 &34.0 &54.0 &34.8 &43.3 &67.5 &47.0 &47.3 &69.5 &57.4 &42 \\ 
Mask R-CNN &ResNet50 & 41.7 &61.4 &45.4 &45.9 &68.0 &50.8 &48.0 &70.5 &58.0 &44  \\ 
RetinaNet &ResNet50 &43.1 &60.5 &46.6 &43.5 &67.1 &48.4 &47.0 &69.5 &56.2 &38 \\ 
DetectoRS &ResNet50 &43.9 &62.9 &47.3 &49.9 &70.7 &53.8 &$\textbf{51.8}$ &71.8 &62.2 &123   \\ 
Swin-T &Swin-Transformer &44.0 &63.9 &48.9 &46.6 &69.8 &51.7 &49.7 &71.0 &60.1 &47 \\   
VitDet &ViT-Base &45.2 &64.1 &50.1 &50.4 &72.4 &55.8 &51.0 &$\textbf{75.1}$ &61.1 &141  \\
VFM-Det (Ours)  &ResNet50/ViT-Base &$\textbf{46.9}$ &$\textbf{66.5}$ &$\textbf{51.6}$ &$\textbf{51.5}$ &$\textbf{75.3}$ &$\textbf{56.6}$ &51.7 &73.7 &$\textbf{63.1}$ & 153 \\   
\hline \toprule [0.5 pt]
\end{tabular} } 
\end{table*}

\begin{table*}[!htp]
\centering 
\caption{Our model loads experimental results of different pre-trained models on three datasets: cityscapes, coco, and ua-detrac. $\dag$ represents pre-trained 100 epochs on the Autobot1M dataset} 
\label{bigmodel} 
\vspace{5pt} 
\begin{tabular}{c|c|ccc|ccc|ccc}  
\hline \toprule [0.5 pt]
\multirow{2}{*}{\raggedright \textbf{Method}} 
&\multirow{2}{*}{\raggedright \textbf{Pre-train Modality}} &\multicolumn{3}{c|}{\textbf{Cityscapes}}  &\multicolumn{3}{c|}{\textbf{COCO}} &\multicolumn{3}{c}{\textbf{UA-DETRAC}} \\
& &$AP_{[0.5:0.95]}$  &$AP_{0.5}$ & $AP_{0.75}$ &$AP_{[0.5:0.95]}$  &$AP_{0.5}$ & $AP_{0.75}$ 
&$AP_{[0.5:0.95]}$  &$AP_{0.5}$ & $AP_{0.75}$ \\
\hline
 DINO &Vision &45.0 &65.6 &49.3 &48.5 &71.6 &54.4 &49.7 &72.6 &60.6  \\
 IBOT &Vision &44.1 &64.8 &48.3 &50.0 &73.5 &55.2 &51.4 &73.1 &63.0  \\
 MoCov3 &Vision &41.8 &60.8 &46.5 &47.5 &69.2 &52.7 &48.5 &71.6 &60.1  \\
 MAE &Vision &42.4 &61.8 &46.4 &47.6 &69.1 &52.9 &49.6 &71.6 &60.6  \\
 MAE$\dag$ &Vision &43.7 &63.0 &47.8 &47.1 &69.2 &52.1 &49.8 &71.9 &60.9  \\
 Ours &Vision and Unmatched Text  &$\textbf{46.9}$ &$\textbf{66.5}$ &$\textbf{51.6}$ &$\textbf{51.5}$ &$\textbf{75.3}$ &$\textbf{56.6}$ &$\textbf{51.7}$ &$\textbf{73.7}$ &$\textbf{63.1}$   \\
\hline \toprule [0.5 pt]
\end{tabular}
\end{table*}

\subsection{Implementation Details} \label{impleDetails} 
In our training phase, the learning rate is set as 0.02, the momentum is 0.9, and the weight decay is 0.0001. The SGD is selected as the optimizer to train our model. The batch size is 2 and training for a total of 26 epochs on every dataset.  All the experiments are implemented using Python based on the deep learning toolkit PyTorch~\cite{paszke2019pytorch}. A server with RTX3090 GPU is used for the training. 

In this work, the attribute head in the \textit{VAtt2Vec} module was pre-trained using the CompCars dataset. To better adapt to the attribute recognition task, we reconstructed the annotated attribute information in the CompCars dataset. Specifically, we modified the maximum speed attribute in the CompCars dataset from its original numerical form to five ranges, i.e., \textit{unknown}, \textit{less than 150}, \textit{greater than 150 and less than 200}, \textit{greater than 200 and less than 250}, and \textit{greater than 250}. Furthermore, based on the classification standards for car engine displacement, we transformed the displacement attribute from its original numerical form to four categories: unknown, small, middle, and large. As a result, our reconstructed dataset comprises six attribute groups, 47 labels, and 44,481 images. A detailed list of attributes is as shown in Table~\ref{attributes}. The attribute head in our \textit{VAtt2Vec} module was pre-trained on the organized CompCars dataset for 20 epochs.


\begin{table}[!htp]
\centering 
\caption{List of Vehicle Attributes} 
\label{attributes} 
\vspace{5pt} 
\begin{tabular}{c|c}  
\hline \toprule [0.5 pt]

 Color & \makecell[c]{
 Black, White, Red \\
 Blue, Greed, Brown, Cyan \\
 Yellow, Gold, Silvery, Grey 
 }  \\ \hline
 Model & \makecell[c]{
 MPV, SUV, Sedan, Hatchback \\ 
 Minibus, Fastback, Estate \\
 Pickup, Hardtop Convertible\\
 Sports, Crossover, Convertible
 }  \\  \hline
 Displacement & \makecell[c]{
 Unkown, Small, Medium, Large  
 }  \\  \hline
 Top speed & \makecell[c]{
 Unkown, Less than 150 \\
 Greater than 150 and less than 200 \\
 Greater than 200 and less than 250 \\
 Greater than 250
 }   \\ \hline
 Number of doors & \makecell[c]{
 Unkown, Two, Three, Four, Five
 }   \\ \hline
 Number of seats & \makecell[c]{
 Unkown, Two, Three, Four, Five \\
 Six, Seven, Eight, Nine, Fifteen \\
 }  \\
\hline \toprule [0.5 pt]
\end{tabular}
\end{table}

\subsection{Compare with SOTA Detectors} \label{citySOTA}

In this experiment, we validated the effectiveness of our proposed VFM-Det model through three vehicle object detection datasets. We compared our baseline model and other advanced object detection models on each dataset. Additionally, we also compared the VehicleMAE with other general pre-training models on the VFM-Det model.


\noindent
$\bullet$ \textbf{Results on Cityscapes datasets.} 
We compared a comparison with other detection algorithms on the Cityscapes dataset, as shown in Table~\ref{detection}, our method VFM-Det, achieved the best performance, reaching $46.9\%$, $66.5\%$, and $51.6\%$ on the $AP_{[0.5:0.95]}$, $AP_{0.5}$, and $AP_{0.75}$ metrics, respectively. Compared to our baseline method Mask R-CNN, we improved by $5.2\%$, $5.1\%$, and $6.2\%$ on the three evaluation metrics, respectively. Compared to VitDet, we improved by $1.7\%$, $2.4\%$, and $ 1.5\%$ on the three evaluation metrics, respectively. Compared with other unsupervised pre-training models, the VehicleMAE we used achieved the best performance. As shown in Table~\ref{bigmodel}, compared to DINO, we improved by $1.9\%$, $0.9\%$, and $2.3\%$ on the three evaluation metrics, respectively. Compared to MAE, we improved by $4.5\%$, $4.7\%$, and $5.2\%$ on the three evaluation metrics, respectively. Additionally, we compared the MAE model pre-trained on our proposed vehicle pre-training dataset, achieving improvements of $3.2\%$, $3.5\%$, and $3.8\%$ on the three evaluation metrics, respectively. 

\noindent
$\bullet$ \textbf{Results on UA-DETRAC datasets.}
We conducted a comparison with other detection algorithms on the UA-DETRAC dataset. As shown in Table~\ref{detection}, our VFM-Det model achieved $51.7\%, 73.7\%$, and $63.1\%$ on the $AP_{[0.5:0.95]}$, $AP_{0.5}$, and $AP_{0.75}$ metrics, respectively. In contrast, our baseline algorithm achieved $48.0\%, 70.5\%$, and $58.0\%$ on these metrics, respectively. Notably, our model outperformed Mask R-CNN by $3.7\%$ on the $AP_{0.5}$ metric. Existing detection algorithms such as VitDet, RetinaNet and DetectoRS fell short of our model in $AP_{0.75}$ metrics. Additionally, we compared our model with other unsupervised pre-training models. As presented in Table~\ref{bigmodel}, our model improved by $3.2\%$ on the $AP_{0.5}$ metric compared to MoCoV3 and by $2.1\%$ compared to MAE. Our model also outperformed other unsupervised pre-training algorithms on other metrics. 


\noindent
$\bullet$ \textbf{Results on COCO2017 datasets.}
We also conducted tests on the vehicle category within the commonly used object detection dataset, COCO2017. As shown in Table~\ref{detection}, our baseline method Mask R-CNN algorithm achieved $45.9\%, 68.0\%$, and $50.8\%$ on the $AP_{[0.5:0.95]}$, $AP_{0.5}$, and $AP_{0.75}$ metrics, respectively. In contrast, our proposed VFM-Det model achieved $51.5\%, 75.3\%$, and $56.6\%$ on these metrics, representing improvements of $5.6\%, 7.3\%$, and $5.8\%$ over Mask R-CNN, respectively. When compared to other detection algorithms, our model outperformed VitDet with improvements of $ 1.1\%, 2.9\%$, and $ 0.8\%$ on the respective metrics. In Table~\ref{bigmodel}, we compared our algorithm with other unsupervised pre-training models on the COCO2017 dataset. Our algorithm improved by $ 4\%$ over MoCoV3 on the $AP_{0.5}$ metrics, and by $ 3.9\%$ over MAE, respectively. These experimental results thoroughly proved the superiority of our model.

\subsection{Ablation Study} \label{ablationStudy}

In this section, to make it easier for readers to understand the contributions of each module in our framework, we conduct comprehensive ablation studies to validate and analyze the effectiveness of the related settings.

\noindent
\textbf{Effects of VehicleMAE Encoder.}
In this paper, we introduce VehicleMAE encoder to enhance the features of the proposal and concatenate it with the original features. As shown in Table \ref{AblationStudy1}, after introducing VehicleMAE encoder, we tested it on the Cityscapes dataset and found that the results improved to $45.0\%, 65.4\%$, and $48.9\%$ on the three metrics. These experimental results indicate that enhanced proposal features can improve detection performance.

\begin{table}[!htp]
\centering 
\caption{Ablation Study on VehicleMAE Encoder and VAtt2Vec Modules.} 
\label{AblationStudy1} 
\vspace{5pt} 
\begin{tabular}{c|ccc}  
\hline \toprule [0.5 pt]
\textbf{Method} &\textbf{$AP_{[0.5:0.95]}$} 
&\textbf{$AP_{0.5}$} & \textbf{$AP_{0.75}$} \\
\hline
 Baseline &41.7 &61.4 &45.4  \\
 +VehicleMAE Encoder &45.0 &65.4 &48.9  \\
 +VAtt2Vec &$\textbf{46.9}$ &$\textbf{66.5}$ &$\textbf{51.6}$  \\
\hline \toprule [0.5 pt]
\end{tabular}
\end{table}

\noindent
\textbf{Effects of VAtt2Vec.}
In the VAtt2Vec module, we introduce the $L_{va}$ loss function. As shown in Table \ref{AblationStudy1}, based on the VehicleMAE encoder, the VAtt2Vec module leads to improvements of $1.9\%$, $1.1\%$, and $2.7\%$ on the three metrics when tested on the Cityscapes dataset. When both the VehicleMAE encoder and VAtt2Vec modules are introduced, the results are further improved to $46.9\%, 66.5\%$, and $51.6\%$. The effectiveness of our VAtt2Vec module can be demonstrated by the experimental results. Since the input features of the attribute head originate from the VehicleMAE encoder, we have not conducted ablation studies specifically targeting the VAtt2Vec alone.

\noindent
\textbf{Effects of the number of learnable tokens.}
Given the large number of proposals generated during the training process, we fixed the parameters of the VehicleMAE encoder during model training to enhance computational efficiency and conserve resources. This action not only effectively reduced computational complexity but also preserved the crucial features and knowledge learned by the large model during previous training. However, fixing the VehicleMAE encoder may result in sub-optimal outcomes, because of discrepancies in the pre-training and detection datasets. Therefore, we introduced a certain number of learnable tokens to enable the pre-trained VehicleMAE encoder to better adapt to new data and tasks. We delved into the influence of the number of introduced learnable tokens on model performance. As shown in Table \ref{AblationStudy2}, when the number of learnable tokens was 4, the respective indicators were $44.8\%, 64.5\%$, and $50.6\%$. When the number increased to 8, each indicator improved by $2.1\%, 2.0\%$, and $1.0\%$, respectively. However, when the number further increased to 12, each indicator decreased by $2.4\%, 2.8\%$, and $1.9\%$, respectively. Based on these experimental results, we decided to introduce 8 learnable tokens in the VehicleMAE encoder. These comparative experiments fully demonstrated the effectiveness of introducing learnable tokens.

\begin{table}[!htp]
\centering 
\caption{Ablation study on the number of learnable tokens.} 
\label{AblationStudy2} 
\vspace{5pt} 
\begin{tabular}{c|ccc}  
\hline \toprule [0.5 pt]
\textbf{Number} &\textbf{$AP_{[0.5:0.95]}$} 
&\textbf{$AP_{0.5}$} & \textbf{$AP_{0.75}$} \\
\hline 
 4 &44.8 &64.5 &50.6  \\
 8 &$\textbf{46.9}$ &$\textbf{66.5}$ &$\textbf{51.6}$  \\
 12 &44.5 &63.7 &49.7  \\
\hline \toprule [0.5 pt]
\end{tabular}
\end{table}

\begin{figure*}
    \centering
    \includegraphics[width=7in]{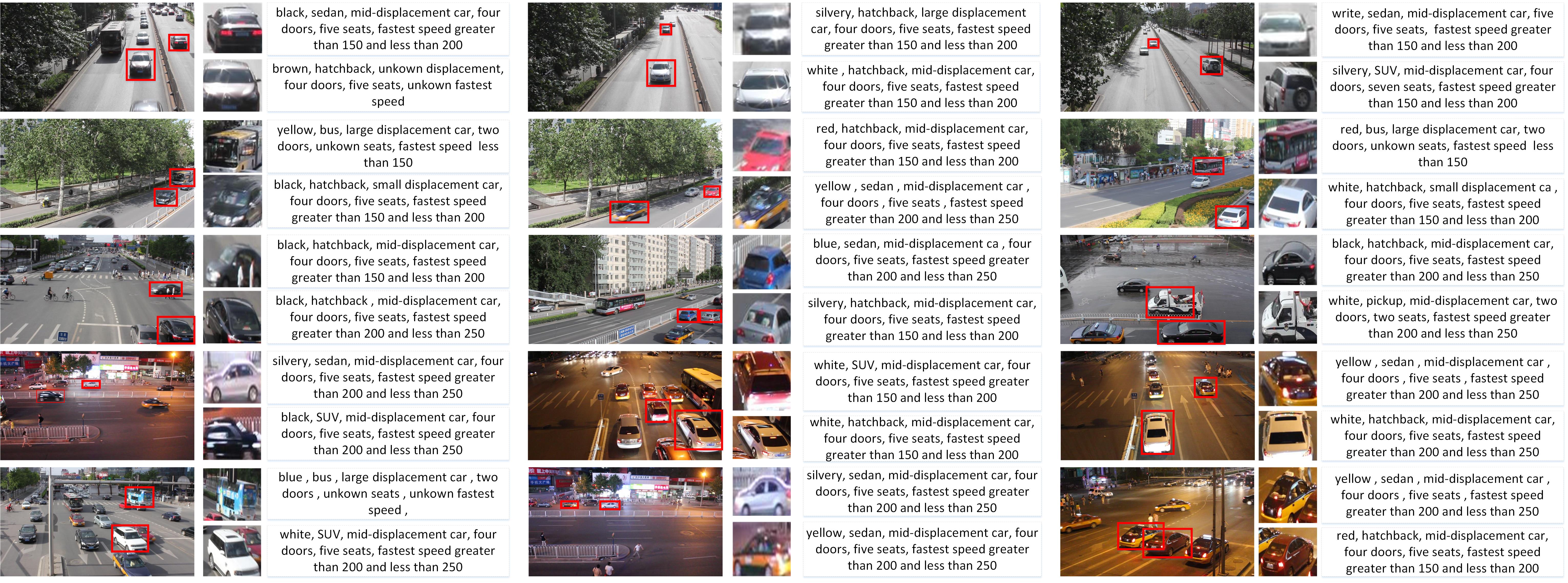}
    \caption{Visualization of the recognized vehicle attributes using the VAtt2Vec module.}
    \label{fig:Visualization2} 
\end{figure*}

\begin{figure*}
    \centering
    \includegraphics[width=7in]{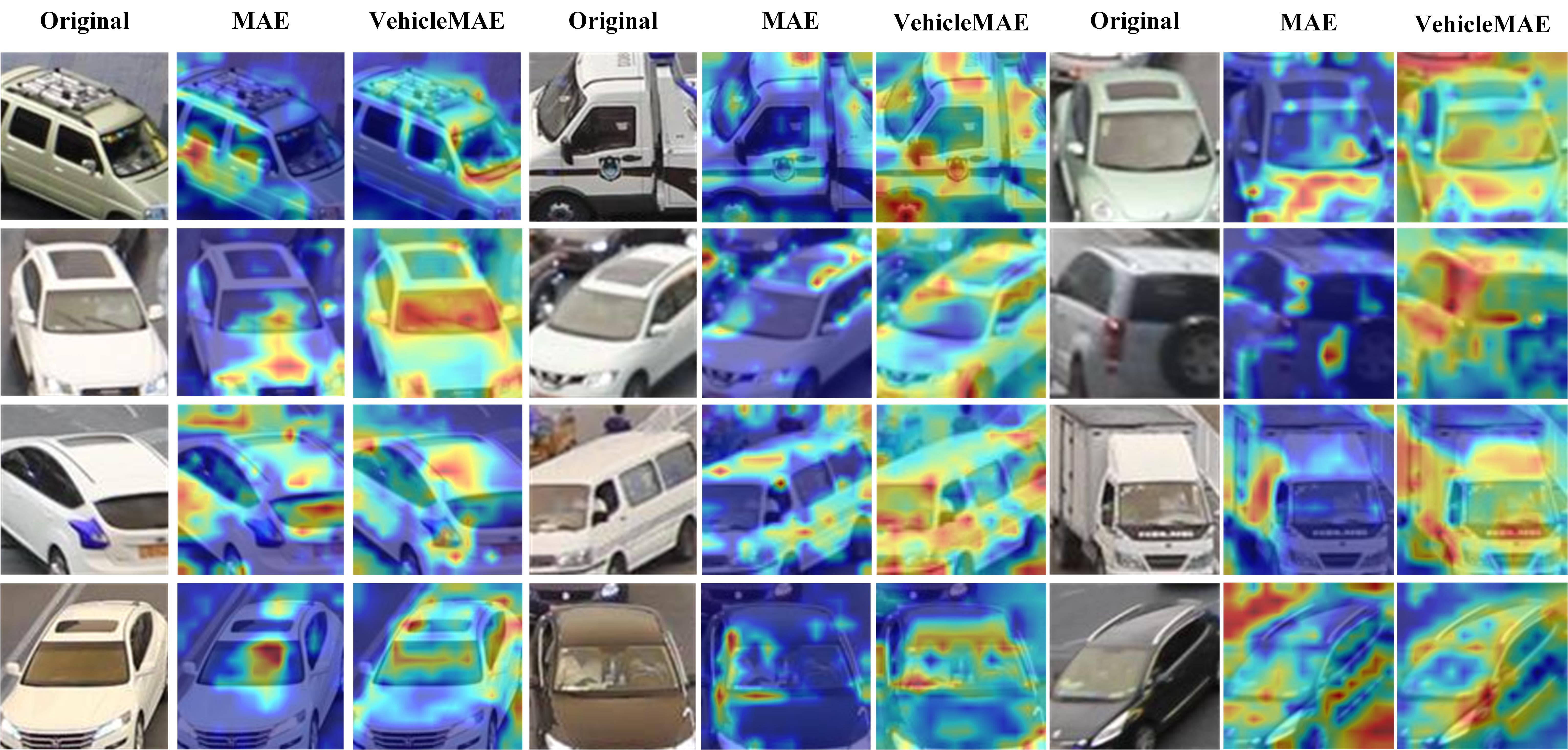}
    \caption{Visualization of attention maps from VehicleMAE encoder.}
    \label{fig:Visualization3} 
\end{figure*}

\noindent
\textbf{Effects of different attribute encoders.}
In this paper, for each candidate region image, a set of corresponding vehicle attributes is predicted. We utilize a large language model to extract features from these attribute information, and then fuse the attribute features into a unified textual representation through a GRU~\cite{cho2014GRU} module. The model is optimized by computing the cosine similarity loss between the textual and visual features. Therefore, the capability of the chosen large language model will directly impact the performance of the detector. We compared five large language models: T5~\cite{raffel2020exploring}, CLIP~\cite{radford2021learning}, BERT~\cite{kenton2019bert}, ALBERT~\cite{lan2019albert}, and MPNet~\cite{song2020mpnet}, in the Table~\ref{AblationStudy3}. Among them, T5~\cite{raffel2020exploring} achieves optimal performance in the $AP_{[0.5:0.95]}$ and $AP_{0.5}$ metrics, with values of $46.9\%$ and $66.5\%$, respectively. However, it falls slightly behind CLIP~\cite{radford2021learning} by $0.4\%$ in the $AP_{0.95}$ metric. Consequently, we choose to use the large language model T5~\cite{raffel2020exploring} as the text encoder.

\begin{table}
\centering 
\caption{Results of different attribute encoders.} 
\label{AblationStudy3} 
\vspace{5pt} 
\begin{tabular}{c|ccc}  
\hline \toprule [0.5 pt]
\textbf{LLM} &\textbf{$AP_{[0.5:0.95]}$} 
&\textbf{$AP_{0.5}$} & \textbf{$AP_{0.75}$} \\
\hline  
BERT~\cite{kenton2019bert} &45.2 &63.1 &51.2  \\
ALBERT~\cite{lan2019albert} &45.9 &64.2 &50.6  \\
MPNet~\cite{song2020mpnet} &45.3 &64.5 &51.4   \\
CLIP~\cite{radford2021learning} &46.5 &66.1 &$\textbf{52.0}$ \\
T5~\cite{raffel2020exploring} &$\textbf{46.9}$ &$\textbf{66.5}$ &51.6  \\
\hline \toprule [0.5 pt]
\end{tabular}
\end{table}

\noindent
\textbf{Effects of different feature fusion strategies.}
In this paper, we retain the image features extracted by the original Mask R-CNN through ResNet50, while simultaneously inputting the candidate region images generated by the RPN into the pre-trained large vehicle model, VehicleMAE, for feature extraction through its encoder. Finally, these two sets of features are fused for detection. Therefore, the method of feature fusion is crucial to the model's performance. We compare three feature fusion strategies: concatenate, weighted fusion, and linear fusion. The effectiveness evaluation results are summarized in Table~\ref{AblationStudy4}. Concatenating the two features achieves $46.9\%, 66.5\%$, and $51.6\%$. Compared to weighted fusion, concatenate operation improves the metrics by $2\%, 3.2\%$, and $2\%$, and compared to linear fusion, it improves by $1\%, 2.3\%$, and $0.7\%$. Consequently, the concatenate operation is chosen as the feature fusion method in this paper.

\begin{table}[!htp]
\centering 
\caption{Results of different vision feature fusion strategies.} 
\label{AblationStudy4} 
\vspace{5pt} 
\begin{tabular}{c|ccc}  
\hline \toprule [0.5 pt]
\textbf{Fusion Strategy} &\textbf{$AP_{[0.5:0.95]}$} 
&\textbf{$AP_{0.5}$} & \textbf{$AP_{0.75}$} \\
\hline  
Weighted Fusion &44.9 &63.2 &49.6  \\
Linear Fusion &45.9 &64.2 &50.9  \\
Concatenate &$\textbf{46.9}$ &$\textbf{66.5}$ &$\textbf{51.6}$  \\
\hline \toprule [0.5 pt]
\end{tabular}
\end{table}

\noindent
\textbf{Effects of different usage methods of attribute vectors.}
Utilizing the attribute heads pre-trained on the CompCars dataset, we conducted attribute prediction for each proposal, obtaining a set of attribute features encoded by T5. Subsequently, these features were fused through a GRU module to derive the textual features of the proposal. We compared two distinct methods of utilizing the attribute vectors, as summarized in Table~\ref{AblationStudy5}. When the textual features were directly concatenated with the image features and then fed into the classification and regression heads, the results were $45.9\%, 64.6\%$, and $50.3\%$, respectively. Alternatively, introducing a cosine similarity loss for image-text contrastive learning yielded results of $46.9\%, 66.5\%$, and $51.6\%$. Compared to the concatenation approach, the contrastive learning method improved the metrics by $1\%, 1.9\%$, and $1.3\%$, respectively. We believe that the primary reason for this improvement lies in the ability of contrastive learning to reduce redundant information and noise while leveraging the complementary between features. Additionally, there exists a domain gap between the CompCars dataset and the proposals, and the contrastive learning approach effectively mitigates this impact compared to concatenation. Consequently, our method adopts the contrastive learning approach.

\begin{table}[!htp]
\centering 
\caption{Results of different usage methods of attribute vectors. $w/o$ denotes without the following item.} 
\label{AblationStudy5} 
\vspace{5pt} 
\begin{tabular}{c|ccc}  
\hline \toprule [0.5 pt]
\textbf{mode} &\textbf{$AP_{[0.5:0.95]}$} 
&\textbf{$AP_{0.5}$} & \textbf{$AP_{0.75}$} \\
\hline  
$w/o$ Attributes &45.0 &65.4 &48.9  \\
Concatenate &45.9 &64.6 &50.3  \\
GRU &$\textbf{46.9}$ &$\textbf{66.5}$ &$\textbf{51.6}$  \\ 
\hline \toprule [0.5 pt]
\end{tabular}
\end{table}

\noindent
\textbf{Effects of different contrastive learning loss functions.~}
In this paper, to alleviate the semantic gap between visual features and semantic categories, we propose to align visual features with a unified attribute representation through the idea of contrastive learning. We compare two contrastive learning loss functions in Table~\ref{AblationStudy6}, namely CLIP~\cite{radford2021learning} loss and cosine embedding loss. From the results, cosine embedding loss achieves optimal performance in the three metrics, with values of $46.9\%$, $66.5\%$ and $51.6\%$, respectively. However, using cross-entropy loss can lead to a decrease in the $AP_{[0.5:0.95]}$ and $AP_{0.5}$ metrics. We believe this is due to the presence of many proposals with the same target, whereas the cross-entropy loss treats all other proposals within a batch as negative samples. Consequently, we choose to use the cosine embedding loss.

\begin{table}[!htp]
\centering 
\caption{Results of different contrastive learning loss functions. $w/o$ denotes without the following item.} 
\label{AblationStudy6} 
\vspace{5pt} 
\begin{tabular}{c|ccc}  
\hline \toprule [0.5 pt]
\textbf{mode} &\textbf{$AP_{[0.5:0.95]}$} 
&\textbf{$AP_{0.5}$} & \textbf{$AP_{0.75}$} \\
\hline 
$w/o$ Contrastive Loss &45.0 &65.4 &48.9  \\
$CLIP$(Cross-entropy Loss) &44.5 &62.6 &50.7  \\
$L_{va}$(Cosine Embedding Loss) &$\textbf{46.9}$ &$\textbf{66.5}$ &$\textbf{51.6}$  \\ 
\hline \toprule [0.5 pt]
\end{tabular}
\end{table}

\noindent
\textbf{Effects of different fine-tuning methods.}
In our proposed VAtt2Vec module, we use a large language model T5 to obtain Vehicle Attributes embeddings. In the paper, we try to use different Parameter-efficient fine-tuning (PEFT) strategies for large language models. The experimental results are shown in the Table~\ref{AblationStudy7}. When without fine-tuning, the model achieves the best performance on both $AP_{[0.5:0.95]}$ and $AP_{0.5}$ metrics. Conversely, fine-tuning the large language model can actually harm the performance of the model. Therefore, we choose not to fine-tune the large language model in our model.

\begin{table}[!htp]
\centering 
\caption{Results of different fine-tuning method. $w/o$ denotes without the following item.}
\label{AblationStudy7} 
\vspace{5pt} 
\begin{tabular}{c|ccc}  
\hline \toprule [0.5 pt]
\textbf{method} &\textbf{$AP_{[0.5:0.95]}$} 
&\textbf{$AP_{0.5}$} & \textbf{$AP_{0.75}$} \\
\hline 
$w/o$ Fine-tuning &$\textbf{46.9}$ &$\textbf{66.5}$ &51.6 \\
Prompt &45.3 &64.6 &50.4  \\
Lora &44.2 &63.3 &49.0  \\ 
Prefix &46.1 &64.4 &$\textbf{52.0}$ \\
\hline \toprule [0.5 pt]
\end{tabular}
\end{table}

\subsection{Visualization} \label{visualization}

In this section, we visualize the detection results of our model, VFM-Det, the attribute results of the proposals detected by the VAtt2Vec module, and the feature maps of the proposals processed by the VehicleMAE backbone.

We present the detection results on road images, where the green boxes represent the detection results of our proposed algorithm, the white boxes indicate the detection results of Mask R-CNN, and the red boxes represent the ground truth. As shown in Fig.~\ref{fig:Visualization1}, our approach can accurately detect vehicle targets. In Fig.~\ref{fig:Visualization2}, we present the detection results of the VAtt2Vec module for proposal attributes. Additionally, as depicted in Fig.~\ref{fig:Visualization3}, we employed GradCAM\footnote{\url{https://mmpretrain.readthedocs.io/en/latest/useful_tools/cam_visualization.html}} to visualize the attention maps of the last Transformer block in the VehicleMAE encoder. It can be observed that the attention is primarily focused on the vehicle targets, indicating that our VehicleMAE encoder is capable of extracting more effective proposal features.

\subsection{Limitation Analysis} \label{subsec:limitation}
Based on the above experiments, we found that introducing the pre-trained vehicle-specific foundation model VehicleMAE, significantly enhances the detection performance for vehicles. However, due to the large number of proposals, even though we fixed the parameters of the VehicleMAE encoder, it still introduced more computational overhead. Therefore, further reducing the complexity of our detector is an important research direction~\cite{wang2024SSMSurvey}. On the other hand, the attribute prediction head is trained on a small dataset which may limit the performance of attribute prediction. This module also makes our detector unable to be optimized end-to-end. 
We will address these two issues in our future work.

\section{Conclusion}  \label{sec::Conclusion}
In this paper, we propose a novel vehicle detection paradigm, named VFM-Det, by extending the region proposal-based detectors based on pre-trained foundation vision and language models. For the input image, we first feed it into a ResNet50 backbone network to obtain image features and generate a set of proposals through a region proposal network. Subsequently, we crop these proposals from the image and utilize the VehicleMAE encoder, which is pre-trained on a large-scale vehicle image dataset, to extract features from the proposals, thus enhancing the raw features. More importantly, we introduce a novel VAtt2Vec module, which predicts the semantic attributes of the vehicle for these proposals based on the features extracted by the VehicleMAE encoder. These attributes are then converted into a unified feature vector, and the model is optimized by computing similarity constraints with visual features. We evaluate and compare our VFM-Det on three vehicle detection datasets, and extensive experiments thoroughly proved the effectiveness and superiority of our proposed vehicle detector.


\bibliographystyle{IEEEtran}
\bibliography{reference}

\vspace{-0.5cm}
\begin{IEEEbiography} [{\includegraphics[width=1in,height=1.25in,clip,keepaspectratio]{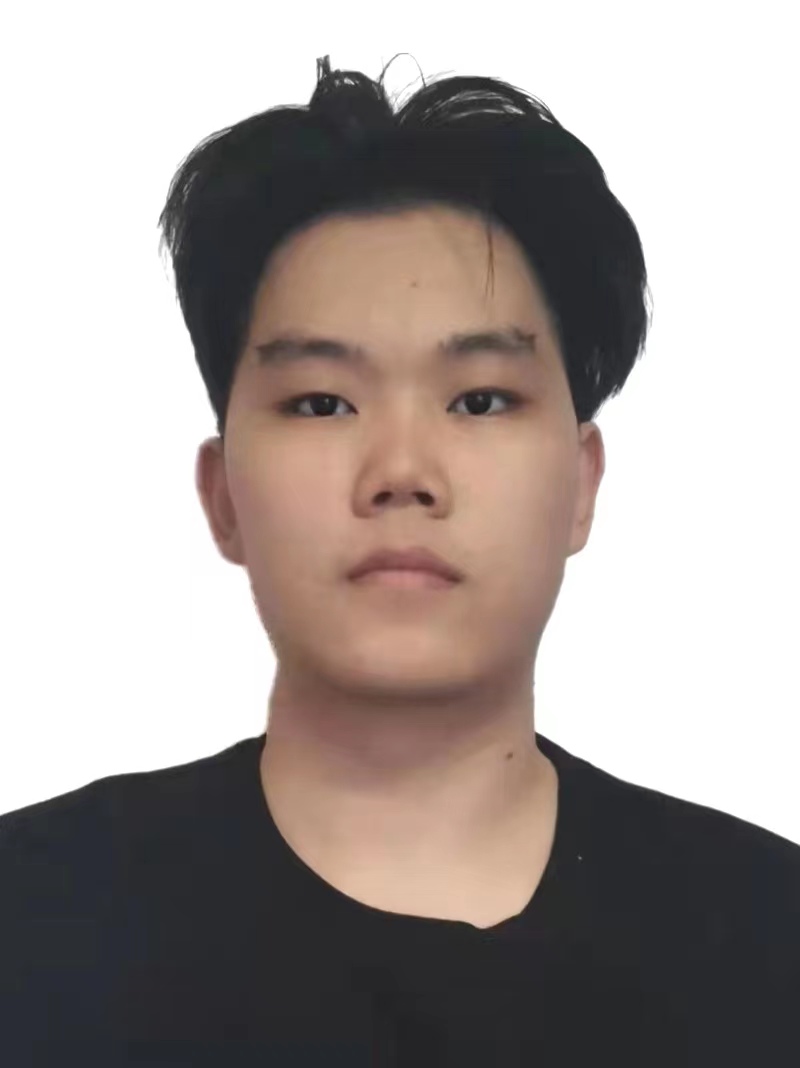}}]
{Wentao Wu} received the B.S. degree from Jiangxi University of Science and Technology, Nanchang, China, in 2022. He is currently working toward the M.S degree with the school of Artificial Intelligence, Anhui University. His research interests include computer vision and deep learning.
\end{IEEEbiography}

\vspace{-0.5cm}
\begin{IEEEbiography} [{\includegraphics[width=1in,height=1.25in,clip,keepaspectratio]{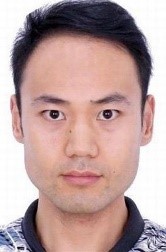}}]
{Fanghua Hong} doctoral candidate at the School of Electronic Information Engineering, Anhui University, specializing in circuits and systems. His primary research area is multimodal visual tracking in the field of computer vision. In 2019, he was appointed as a senior software engineer in computer science and has spent many years conducting video-tracking research in the public affairs management department.
\end{IEEEbiography}

\vspace{-0.5cm}
\begin{IEEEbiography} [{\includegraphics[width=1in,height=1.25in,clip,keepaspectratio]{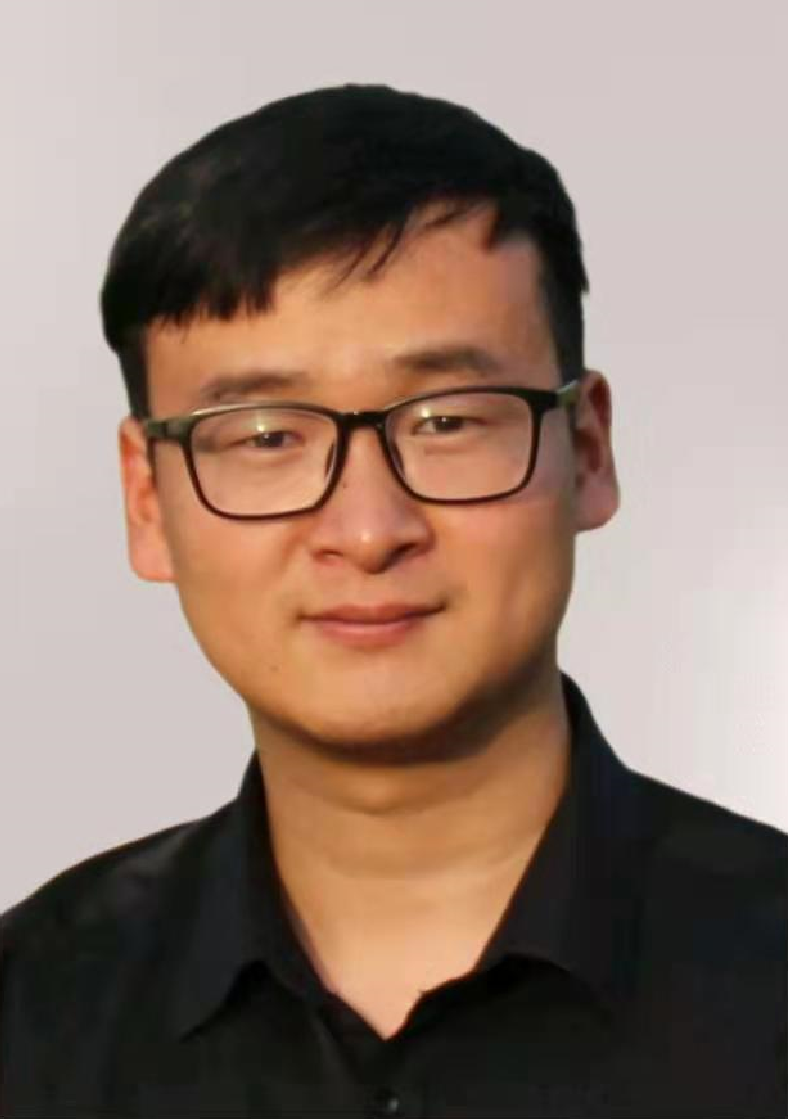}}]
{Xiao Wang} (Member, IEEE) received the B.S. degree from West Anhui University, Luan, China, in 2013. He received the Ph.D. degree in computer science from Anhui University, Hefei, China, in 2019. From 2015 and 2016, he was a visiting student at the School of Data and Computer Science, Sun Yat-sen University, Guangzhou, China. He also has a visit to UBTECH Sydney Artificial Intelligence Centre, the Faculty of Engineering, the University of Sydney, in 2019. He finished the postdoc research at Peng Cheng Laboratory, Shenzhen, China, from April 2020 to April 2022. He is now an Associate Professor at the School of Computer Science and Technology, Anhui University, Hefei, China. His current research interests are mainly Computer Vision, Event-based Vision, Machine Learning, and Pattern Recognition. He serves as a reviewer for a number of journals and conferences, such as IEEE TCSVT, TNNLS, TIP, IJCV, CVIU, PR, CVPR, ICCV, AAAI, ECCV, ACCV, ACM-MM, and WACV. \textbf{Homepage}: \url{https://wangxiao5791509.github.io/} 
\end{IEEEbiography}

\vspace{-0.5cm}
\begin{IEEEbiography} [{\includegraphics[width=1in,height=1.25in,clip,keepaspectratio]{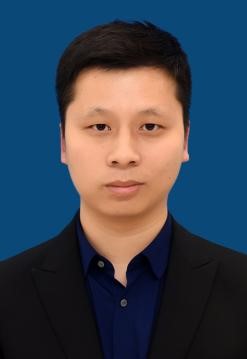}}]
{Chenglong Li} received the M.S. and Ph.D. degrees from the School of Computer Science and Technology,Anhui University, Hefei, China, in 2013 and 2016, respectively. From 2014 to 2015, he was a Visiting Student at the School of Artificial Intelligence, Sun Yat-sen University, Guangzhou, China. He was a Post-Doctoral Research Fellow with the Center for Research on Intelligent Perception and Computing (CRIPAC), National Laboratory of Pattern Recognition (NLPR), Institute of Automation, Chinese Academy of Sciences (CASIA), Beijing, China. He is currently a Professor and a Ph.D. Supervisor with the School of Artificial Intelligence, Anhui University. His research interests include computer vision and deep learning
\end{IEEEbiography}

\vspace{-0.5cm}
\begin{IEEEbiography} [{\includegraphics[width=1in,height=1.25in,clip,keepaspectratio]{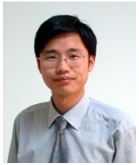}}]
{Jin Tang} received the B.Eng. Degree in automation and the Ph.D. degree in computer science from Anhui University, Hefei, China, in 1999 and 2007, respectively. He is currently a Professor at the School of Computer Science and Technology, Anhui University. His current research interests include computer vision, pattern recognition, and machine learning.
\end{IEEEbiography}

\end{document}